\documentclass[11pt,a4paper]{article}
\usepackage[hyperref]{acl2020}
\usepackage{times}
\usepackage{latexsym}

\usepackage{graphicx}

\usepackage{microtype}

\aclfinalcopy

\title{Enhancing Pre-trained Language Model with Lexical Simplification}

\author{Rongzhou Bao \\
  Affiliation / Address line 2 \\
  Affiliation / Address line 3 \\
  \texttt{rongzhou.bao@outlook.com} \\\And
  Jiayi Wang \\
  \texttt{email@domain} \\\And
  Zhuosheng Zhang \\

  \texttt{email@domain} \\\And
  Hai Zhao}
 
 \author{Rongzhou Bao\textsuperscript{1,2,3}, Jiayi Wang\textsuperscript{1,2,3}, Zhuosheng Zhang\textsuperscript{1,2,3,}, Hai Zhao\textsuperscript{1,2,3}\\
\textsuperscript{1} Department of Computer Science and Engineering, Shanghai Jiao Tong University\\
\textsuperscript{2} Key Laboratory of Shanghai Education Commission for Intelligent Interaction\\
and Cognitive Engineering, Shanghai Jiao Tong University, Shanghai, China\\
\textsuperscript{3}MoE Key Lab of Artificial Intelligence, AI Institute, Shanghai Jiao Tong University, Shanghai, China\\
\texttt{rongzhou.bao@outlook.com}\\
\texttt{\{wangjiayi\_102\_23,zhangzs\}@sjtu.edu.cn,zhaohai@cs.sjtu.edu.cn}\\
}

\date{}

\begin{document}
\maketitle
\begin{abstract}
For both human readers and pre-trained language models (PrLMs), lexical diversity may lead to confusion and inaccuracy when understanding the underlying semantic meanings of given sentences. By substituting complex words with simple alternatives, lexical simplification (LS) is a recognized method to reduce such lexical diversity, and therefore to improve the understandability of sentences. In this paper, we leverage LS and propose a novel approach which can effectively improve the performance of PrLMs in text classification. A rule-based simplification process is applied to a given sentence. PrLMs are encouraged to predict the real label of the given sentence with auxiliary inputs from the simplified version. Using strong PrLMs (BERT and ELECTRA) as baselines, our approach can still further improve the performance in various text classification tasks.
\end{abstract}

\section{Introduction}

Pre-trained language models (PrLMs) such as BERT \cite{devlin:bert}, RoBERTa \cite{liu2019roberta}, and ELECTRA \cite{clark2020electra} have led to strong performance gains in downstream natural language understanding (NLU) tasks, including text classification. However, \cite{li-etal-2020-bert-attack,jin2019bert} demonstrate that it only takes a few simple synonym replacements to mislead the prediction of PrLMs on various text classification tasks. Such result indicates that lexical diversity can pose a negative impact on the accuracy of semantic meaning understanding for PrLMs. 

In order to reduce lexical diversity, previous works have proposed some approaches for lexical simplification (LS) \cite{gooding-kochmar-2019-recursive,qiang2019BERTLS}. By substituting complex words with their simpler alternatives in original sentences, LS can generate a simplified sentence version, which is much easier to understand for human readers. Inspired by these studies, we leverage LS as a paraphrasing tool to enhance the prediction accuracy of PrLMs in text classification tasks. 

A well-designed LS rule customized to neural network (e.g. PrLM) is crucial for our overall approach. However, existing LS methods are not suitable for PrLMs. Current methods \cite{gooding-kochmar-2019-recursive,qiang2019BERTLS} are mainly for human readers to simplify reading process, but not for neural network to improve prediction accuracy. Furthermore, current LS methods are very time-consuming. This is because they apply large pre-trained neural networks to detect and replace the complex words in a recursive way\cite{qiang2019BERTLS}. Therefore, we design a lexical simplification method based on lemmatization and rare word replacement (abbreviated as LRLS), which is more effective and serves better to our purpose, to generate simplified version of give sentence.

In order to better accommodate the LRLS lexical simplification method with PrLMs and improve the overall performance, an auxiliary framework is designed and executed. The simplified sentence generated by LRLS serves as an auxiliary input in both training and inference phase of PrLMs. In this way, PrLMs are able to make the right decision based on both the original sentence and the simplified perspective. Thus, the challenge posed by lexical diversity in text classification can be significantly reduced.

\begin{table*}[!htp]
    \centering
    {
        \begin{tabular}{l l l l l l l}
        \hline \textbf{Model} & \textbf{SST-2} & \textbf{MR}& \textbf{CR}& \textbf{SUBJ}&\textbf{AG}&\textbf{Avg} \  \\ \hline
        BERT$_{BASE}$ & 92.4 &86.1&90.0&97.3&94.2 & 92.0 \\
        +LS  & 93.5(+1.1)  & 88.1(+2.0)&90.8(+0.8) &98.0(+0.7)&95.0(+0.8)& 93.1(+1.1)\\
        \hline
        ELECTRA$_{LARGE}$ & 96.7  &90.0&94.3&97.4&94.6&94.6 \\
        +LS & \textbf{97.5}(+0.8)   & \textbf{91.4}(+1.4)&\textbf{94.5}(+0.2) &\textbf{98.1}(+0.7)&\textbf{95.3}(+0.7)&\textbf{95.3}(+0.7)\\
        \hline
        \end{tabular}
}
\caption{\label{tab:result} performances (\%) across five text
classification tasks for models with and without LS. }
\end{table*}

A series experiments are conducted on various text classification tasks. Empirical results show that our approach can notably improve the performance PrLMs. Meanwhile, ablation studies prove the effectiveness of our LRLS method. Furthermore, we also compare our LRLS method with other paraphrasing method used in data augmentation, such as randomly replacement of several words by synonyms \cite{wu2019conditional,wei-zou-2019-eda}, back-translation \cite{xie2019unsupervised,edunov2018understanding}, cutoff \cite{shen2020simple}. Analysis result demonstrate that our LRLS method remains the most effective.

\section{Method}
\subsection{LRLS Lexical Simplification Process}
A well-adapted LS process is the essential to our approach. Previous works \cite{li-etal-2020-bert-attack,jin2019bert} show that the prediction of PrLMs would be easily misled by replacing only a few words with their synonyms in the given sentences. By carefully observing the adversarial examples, we find that changing the tense of verbs, changing the singular and plural form of nouns, and replacing words by its less frequent synonyms compose the majority of the adversarial examples. The observation is also confirmed by \cite{mozes2020frequencyguided}.

Inspired by such observation, our LRLS method is developed with two major steps: (1) lemmataization by transforming verbs and nouns into corresponding lemmas, and (2) replacing rare words with theirs more common synonyms. Firstly, we employ Natural Language Toolkit (NLTK) to detect the verbs and nouns in the given sentences, and transform every verb to its infinitive form and every noun to its singular form. Secondly, according to a word frequency list\footnote{https://github.com/hermitdave/FrequencyWords}, we label every word whose frequency is less than a frequency threshold $n_f$ as a rare word in the given sentence. We then use a word embedding from \cite{2016counterfitting}, which is specially curated for locating synonyms, to find the top $n_s$ synonyms of identified rare words with the highest cosine similarity. Each rare word is replaced by its synonym with the highest frequency. A part-of-speech (POS) check is also applied to ensure that all the synonymous candidates hold the same POS as the original words. 
\subsection{Simplified Sentence As Auxiliary Input}
Following \cite{devlin:bert}, the original sentence and its simplified version are combined together in to a single sentence. In our approach, the original and simplified sentences are differentiated in two ways. First, a special separation token ([SEP]) is inserted between the two sentences. Second, a learned segmentation embedding is added to every token which indicates whether it belongs to the original sentence or the simplified sentence. In both training and inference phases, we feed PrLMs the original-simplified sequence as inputs. The rest of implementations remain the same as the original PrLMs.

\begin{figure*}
	\centering
	\includegraphics[width=0.9\textwidth]{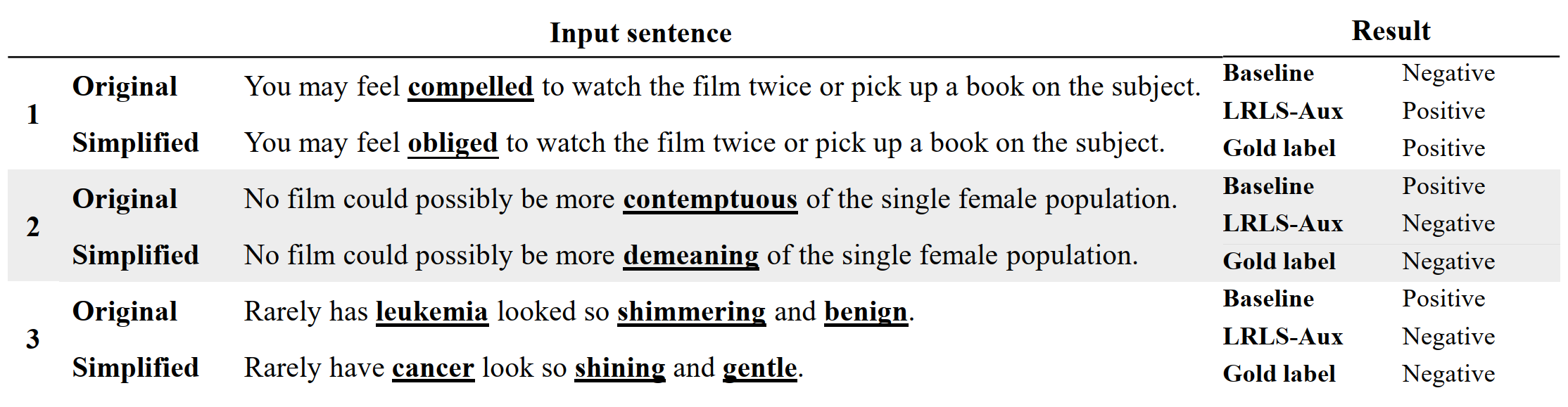}
	\caption{\label{fig:ls_exp}Examples that show how auxiliary inputs from simplified sentences help the PrLMs to mlrake the right prediction. In the result column. \textbf{Baseline} demonstrates the original prediction made by BERT, and \textbf{LRLS-Aux} shows the prediction generated with the auxiliary inputs from simplified sentences.    }
\end{figure*}

\section{Experimental Setup}
\subsection{Benchmark Datasets}
We conduct our experiments on five benchmark text
classification tasks: (1) SST-2: Stanford Sentiment Treebank \cite{socher2013recursive}, (2) CR: customer reviews \cite{10.1145/1014052.1014073,10.5555/2832415.2832429},
(3) SUBJ: subjectivity/objectivity dataset \cite{pang-lee-2004-sentimental}, (4) MR:Movie reviews \cite{pang-lee-2005-seeing}, and (5) AG: AG’s News, classification task with regard to four news topics: World, Sports, Business, and Science.
\subsection{Baseline Models}
We use (1) BERT-base \cite{devlin:bert} with 12 layers, 768 hidden units, 12 heads and 110M parameters, and (2) ELECTRA-large \cite{clark2020electra} with 24 layers, 1024 hidden units,
16 heads and 340M parameters as our baseline PrLMs.

\section{Experiments}
In this section, comprehensive experiments and analysis are conducted. For all the experiments, we average results from three different random seeds.

\subsection{Our Approach Make Gains}
As shown in Table \ref{tab:result}, we run both BERT-base \cite{devlin:bert} and ELECTRA-large \cite{clark2020electra}, with and without LS, across all five datasets. The average gain is 1.1 for BERT-base and 0.7 for ELECTRA-large. As ELECTRA-large is a very strong baseline, the result prove the effectiveness of our approach. As show in Figure \ref{fig:ls_exp}, we select several examples from MR and SST-2 to further illustrate how PrLMs can benefit from the auxiliary input of simplified sentences.

\subsection{Impact of Lexical Simplification Process }

Since our LRLS method is composed of two steps: transformation of verbs and nouns into their lemmas, and replacement of rare words. To investigate the impact of different LS methods, we firstly apply the two steps separately and compare with our LRLS method. We also include BERT-LS \cite{qiang2019BERTLS}, which leverages masking language model of BERT to generate synonym candidates of rare words, for further comparison.

As shown in Table \ref{tab:LS}, the lemma transformation and rare words replacement are both effective, but we can further improve the performance by combining these two methods together. The performance of our method also exceeds that of BERT-LS. Moreover, our method is more than a hundred faster than BERT-LS, since our method is entirely rule-based, while BERT-LS uses a large pre-trained neural network to detect and replace the complex words recursively.

\begin{table}[!htp]
    \centering
    {
        \begin{tabular}{l c c }
        \hline \textbf{Method} & \textbf{MR} & \textbf{SST-2} \\
        \hline
        BERT$_{BASE}$ & 86.4 &92.4  \\
        Lemma & 87.6  & 93.1\\
        RR & 87.7  &92.9 \\
        BERT LS & 87.9  & 93.1\\
        LRLS & 88.1  & 93.5\\
        \hline
        \end{tabular}
}
\caption{\label{tab:LS} Performances (\%) using different LS methods. \textbf{Lemma} represents the transformation of verbs and nouns into their lemma, \textbf{RR} represents the replacement of rare words.}
\end{table}

\subsection{Words Replacement Hyperparameters }

The process of the rare word replacement is controlled by two hyperparameters: $n_f$ and $n_s$. $n_f$ is the frequency threshold under which the word will be labelled as rare word and replaced. The larger the $n_f$, the more words will be replaced. $n_s$ is the number of synonym candidates. The larger the $n_s$, the larger possibility that the rare words will be replaced by more common but less similar candidates. 
In order to investigate the effect of these two hyper-parameters, we change these two hyperparameters separately and conduct experiments on MR and SST-2 to see the impact on the performance. 

As shown in Figure 2, the best performance gain is obtained with middle-sized $n_f$ and $n_s$, which is consistent with our expectation. Because if $n_f$ and $n_s$ is too small the simplified sentence will be almost the same as the original version, on the contrary if $n_f$ and $n_s$ is too large, it may change the underlying meaning of the sentence. 

\begin{figure}
	\centering
	\includegraphics[width=0.5\textwidth]{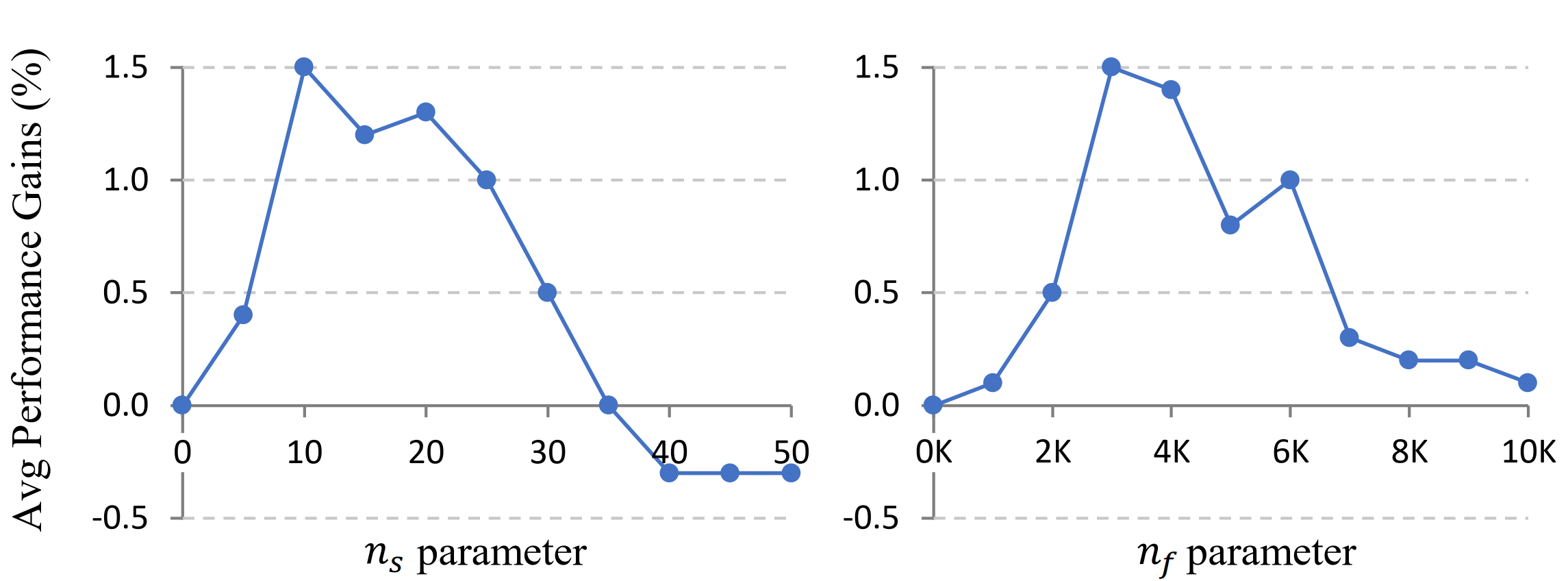}
	\caption{\label{fig:hyperparameter}Average performance gain over MR and SST-2. $n_s$ is the number of synonym candidates and $n_f$ is the threshold of the frequency under which the word will be replaced.}
\end{figure}

\subsection{Alternative Frameworks }
We use simplified sentences as auxiliary inputs to improve the prediction accuracy of PrLMs. However, there are other frameworks to incorporate lexical simplification with PrLMs.

One alternative framework is to feed PrLM only the simplified sentences in both training and inference phases. In this case, prediction is made solely based on simplified versions.

Another framework is to leverage LS as a data augmentation technique. To illustrate, let $D = \{x_i,y_i\}_{i=1 \dots N}$ denote the training dataset. For a given sample $\{x_i,y_i\}$ in the training dataset, we generate an augmented sample by simplifying the sentence $x_i$ to $x'_i$ and preserving the label $y_i$. In this way, we generate an augmented dataset $D' = \{x'_i,y_i\}_{i=1 \dots N}$. PrLMs can thus learn from both the training set $D$ and the augmented set $D'$. 

Experiments are conducted to compared our framework with the two alternative frameworks mentioned above on BERT-base.

\begin{table}[!htp]
    \centering
    {
        \begin{tabular}{l c c }
        \hline \textbf{Method} & \textbf{MR} & \textbf{SST-2} \\ 
        \hline
        BERT$_{BASE}$ & 86.4 &92.4  \\
        LRLS only  & 86.5  & 92.1\\
        LRLS Aug & 87.9  &92.6 \\
        LRLS Aux & 88.1  & 93.5\\
        \hline
        \end{tabular}
}
\caption{\label{framework} Performances (\%) using different frameworks to leverage simplified sentences. \textbf{LRLS only} represents predictions made solely based on simplified sentences, \textbf{LRLS Aug} represents the use of simplified sentences for training data augmentation, \textbf{LRLS Aux} represents using simplified sentences as auxiliary inputs.}
\end{table}

As show in Table \ref{framework}, framework using simplified sentences as the only input (\textbf{LRLS only}) would slightly harm the performance of PrLM. This is because a part of semantic meanings carried by original sentences may be lost during the simplification process. Experiments also show that leveraging lexical simplification for data augmentation (\textbf{LRLS Aug}) is also beneficial for the overall performance. However, this framework would double the training time and the performance is still worse than our framework (\textbf{LRLS Aux}) .

\subsection{Alternative Paraphrasing Methods}
While we leverage LRLS method to paraphrase the original sentence and generate auxiliary inputs for PrLMs, we wonder if other commonly used paraphrasing techniques are effective. 

These paraphrasing methods include (1) random replacement of several words by their synonyms \cite{wu2019conditional,wei-zou-2019-eda}, (2) translating an existing example $x$ in language A into another language B, and then translating it back into A to obtain a paraphrased example $x'$ (back-translation) \cite{xie2019unsupervised,edunov2018understanding} , and (3) randomly delete several words in the sentence (cutoff) \cite{shen2020simple}.

The upper mentioned paraphrasing methods are applied on original sentences respectively to generate auxiliary inputs, and then incorporated into PrLMs. Performance on MR and SST-2 from different paraphrasing methods are compared.

As show in Table \ref{paraphrase}, cutoff would slightly harm the overall performance. This is because it simply randomly deletes several words in the original sentence to generate a paraphrased version, which tends to twist the original semantic meaning and adds noise for predictions. Although back-translation and random replacement can slightly boost the performance of PrLMs, our LRLS method remains the most effective.
\begin{table}[!htp]
    \centering
    {
        \begin{tabular}{l c c }
        \hline \textbf{Method} & \textbf{MR} & \textbf{SST-2} \\ \hline
        BERT$_{BASE}$ & 86.4 &92.4  \\
        +back-translation  & 87.0  & 92.8\\
        +cutoff & 86.3  &91.6 \\
        +random replacement & 87.3  & 92.5\\
        +LRLS & 88.0  & 93.5\\
        \hline
        \end{tabular}
}
\caption{\label{paraphrase} Performances (\%) using different paraphrasing techniques to generate auxiliary inputs. }
\end{table}

\section{Conclusion}
This paper proposes a novel approach that leverages lexical simplification and to reduce lexical diversity and enhance the performance of PrLMs on text classification. Experiments on various text classification tasks demonstrate that our approach consistently improves strong baselines. 

Within the framework, we incorporate a specially designed lexical simplification process based on lemmatization and rare word replacement (LRLS) for better performance. Our comprehensive analysis also show that compared with other paraphrasing techniques used in previous works, LRLS is a more effective paraphrasing method to offer auxiliary information for prediction. 

Furthermore, an effective framework (LRLS Aux) leveraging LRLS as auxiliary information is designed. Unlike data augmentation which only leverages paraphrased information in training phase, LRLS Aux incorporates the information in both training and inference phase and achieves better performance gains. Such framework may shed the light for more future studies.

\newpage
\bibliography{anthology,acl2020}

\begin{thebibliography}{19}
\expandafter\ifx\csname natexlab\endcsname\relax\def\natexlab#1{#1}\fi

\bibitem[{Clark et~al.(2020)Clark, Luong, Le, and Manning}]{clark2020electra}
Kevin Clark, Minh-Thang Luong, Quoc~V. Le, and Christopher~D. Manning. 2020.
\newblock {ELECTRA}: Pre-training text encoders as discriminators rather than
  generators.
\newblock In \emph{ICLR}.

\bibitem[{Devlin et~al.(2018)Devlin, Chang, Lee, and Toutanova}]{devlin:bert}
Jacob Devlin, Ming-Wei Chang, Kenton Lee, and Kristina Toutanova. 2018.
\newblock {BERT}: Pre-training of deep bidirectional transformers for language
  understanding.
\newblock \emph{arXiv preprint arXiv:1810.04805}.

\bibitem[{Edunov et~al.(2018)Edunov, Ott, Auli, and
  Grangier}]{edunov2018understanding}
Sergey Edunov, Myle Ott, Michael Auli, and David Grangier. 2018.
\newblock \href {http://arxiv.org/abs/1808.09381} {Understanding
  back-translation at scale}.

\bibitem[{Gooding and Kochmar(2019)}]{gooding-kochmar-2019-recursive}
Sian Gooding and Ekaterina Kochmar. 2019.
\newblock \href {https://doi.org/10.18653/v1/D19-1491} {Recursive context-aware
  lexical simplification}.
\newblock In \emph{Proceedings of the 2019 Conference on Empirical Methods in
  Natural Language Processing and the 9th International Joint Conference on
  Natural Language Processing (EMNLP-IJCNLP)}, pages 4855--4865, Hong Kong,
  China. Association for Computational Linguistics.

\bibitem[{Hu and Liu(2004)}]{10.1145/1014052.1014073}
Minqing Hu and Bing Liu. 2004.
\newblock \href {https://doi.org/10.1145/1014052.1014073} {Mining and
  summarizing customer reviews}.
\newblock In \emph{Proceedings of the Tenth ACM SIGKDD International Conference
  on Knowledge Discovery and Data Mining}, KDD '04, New York, NY, USA.
  Association for Computing Machinery.

\bibitem[{Jin et~al.(2019)Jin, Jin, Zhou, and Szolovits}]{jin2019bert}
Di~Jin, Zhijing Jin, Joey~Tianyi Zhou, and Peter Szolovits. 2019.
\newblock Is bert really robust? natural language attack on text classification
  and entailment.
\newblock \emph{arXiv preprint arXiv:1907.11932}.

\bibitem[{Li et~al.(2020)Li, Ma, Guo, Xue, and Qiu}]{li-etal-2020-bert-attack}
Linyang Li, Ruotian Ma, Qipeng Guo, Xiangyang Xue, and Xipeng Qiu. 2020.
\newblock {BERT}-{ATTACK}: Adversarial attack against {BERT} using {BERT}.
\newblock In \emph{Proceedings of the 2020 Conference on Empirical Methods in
  Natural Language Processing (EMNLP)}, Online. Association for Computational
  Linguistics.

\bibitem[{Liu et~al.(2015)Liu, Gao, Liu, and Zhang}]{10.5555/2832415.2832429}
Qian Liu, Zhiqiang Gao, Bing Liu, and Yuanlin Zhang. 2015.
\newblock Automated rule selection for aspect extraction in opinion mining.
\newblock In \emph{Proceedings of the 24th International Conference on
  Artificial Intelligence}, IJCAI'15, page 1291–1297. AAAI Press.

\bibitem[{Liu et~al.(2019)Liu, Ott, Goyal, Du, Joshi, Chen, Levy, Lewis,
  Zettlemoyer, and Stoyanov}]{liu2019roberta}
Yinhan Liu, Myle Ott, Naman Goyal, Jingfei Du, Mandar Joshi, Danqi Chen, Omer
  Levy, Mike Lewis, Luke Zettlemoyer, and Veselin Stoyanov. 2019.
\newblock Roberta: A robustly optimized bert pretraining approach.
\newblock \emph{arXiv preprint arXiv:1907.11692}.

\bibitem[{Mozes et~al.(2020)Mozes, Stenetorp, Kleinberg, and
  Griffin}]{mozes2020frequencyguided}
Maximilian Mozes, Pontus Stenetorp, Bennett Kleinberg, and Lewis~D. Griffin.
  2020.
\newblock \href {http://arxiv.org/abs/2004.05887} {Frequency-guided word
  substitutions for detecting textual adversarial examples}.

\bibitem[{Mrkšić et~al.(2016)Mrkšić, Séaghdha, Thomson, Gašić,
  Rojas-Barahona, Su, Vandyke, Wen, and Young}]{2016counterfitting}
Nikola Mrkšić, Diarmuid~Ó Séaghdha, Blaise Thomson, Milica Gašić, Lina
  Rojas-Barahona, Pei-Hao Su, David Vandyke, Tsung-Hsien Wen, and Steve Young.
  2016.
\newblock \href {http://arxiv.org/abs/1603.00892} {Counter-fitting word vectors
  to linguistic constraints}.

\bibitem[{Pang and Lee(2004)}]{pang-lee-2004-sentimental}
Bo~Pang and Lillian Lee. 2004.
\newblock \href {https://doi.org/10.3115/1218955.1218990} {A sentimental
  education: Sentiment analysis using subjectivity summarization based on
  minimum cuts}.
\newblock In \emph{Proceedings of the 42nd Annual Meeting of the Association
  for Computational Linguistics ({ACL}-04)}, pages 271--278, Barcelona, Spain.

\bibitem[{Pang and Lee(2005)}]{pang-lee-2005-seeing}
Bo~Pang and Lillian Lee. 2005.
\newblock \href {https://doi.org/10.3115/1219840.1219855} {Seeing stars:
  Exploiting class relationships for sentiment categorization with respect to
  rating scales}.
\newblock In \emph{Proceedings of the 43rd Annual Meeting of the Association
  for Computational Linguistics ({ACL}{'}05)}, pages 115--124, Ann Arbor,
  Michigan. Association for Computational Linguistics.

\bibitem[{Qiang et~al.(2020)Qiang, Li, Yi, Yuan, and Wu}]{qiang2019BERTLS}
Jipeng Qiang, Yun Li, Zhu Yi, Yunhao Yuan, and Xindong Wu. 2020.
\newblock Lexical simplification with pretrained encoders.
\newblock \emph{AAAI}.

\bibitem[{Shen et~al.(2020)Shen, Zheng, Shen, Qu, and Chen}]{shen2020simple}
Dinghan Shen, Mingzhi Zheng, Yelong Shen, Yanru Qu, and Weizhu Chen. 2020.
\newblock \href {http://arxiv.org/abs/2009.13818} {A simple but tough-to-beat
  data augmentation approach for natural language understanding and
  generation}.

\bibitem[{Socher et~al.(2013)Socher, Perelygin, Wu, Chuang, Manning, Ng, and
  Potts}]{socher2013recursive}
Richard Socher, Alex Perelygin, Jean Wu, Jason Chuang, Christopher~D Manning,
  Andrew Ng, and Christopher Potts. 2013.
\newblock Recursive deep models for semantic compositionality over a sentiment
  treebank.
\newblock In \emph{EMNLP}.

\bibitem[{Wei and Zou(2019)}]{wei-zou-2019-eda}
Jason Wei and Kai Zou. 2019.
\newblock \href {https://www.aclweb.org/anthology/D19-1670} {{EDA}: Easy data
  augmentation techniques for boosting performance on text classification
  tasks}.
\newblock In \emph{Proceedings of the 2019 Conference on Empirical Methods in
  Natural Language Processing and the 9th International Joint Conference on
  Natural Language Processing (EMNLP-IJCNLP)}, pages 6383--6389, Hong Kong,
  China. Association for Computational Linguistics.

\bibitem[{Wu et~al.(2019)Wu, Lv, Zang, Han, and Hu}]{wu2019conditional}
Xing Wu, Shangwen Lv, Liangjun Zang, Jizhong Han, and Songlin Hu. 2019.
\newblock Conditional bert contextual augmentation.
\newblock In \emph{International Conference on Computational Science}, pages
  84--95. Springer.

\bibitem[{Xie et~al.(2019)Xie, Dai, Hovy, Luong, and Le}]{xie2019unsupervised}
Qizhe Xie, Zihang Dai, Eduard Hovy, Minh-Thang Luong, and Quoc~V Le. 2019.
\newblock Unsupervised data augmentation for consistency training.
\newblock \emph{arXiv preprint arXiv:1904.12848}.

\end{thebibliography}
\bibliographystyle{acl_natbib}

\appendix

\end{document}